\documentclass[10pt,journal,compsoc]{IEEEtran}
\usepackage[table,dvipsnames]{xcolor}
\usepackage{tikz}

\ifCLASSOPTIONcompsoc
  \usepackage[nocompress]{cite}
\else
  \usepackage{cite}
\fi

\ifCLASSINFOpdf
\else
\fi

\usepackage{pgfplots}
\pgfplotsset{compat=newest}
\usepackage{booktabs} %
\usepackage{etoolbox,siunitx} %
\usepackage{pgfplotstable} %
\usetikzlibrary{plotmarks} %
\usetikzlibrary{colorbrewer} %
\usepackage{adjustbox}

\usepackage{rotating}
\usepackage{graphicx}
\usepackage{amsmath,amssymb} %
\usepackage{url}

\usepackage{color}
\usepackage{pifont}

\definecolor{olivegreen}{RGB}{0,170,0}
\definecolor{darkred}{RGB}{220,100,10}
\definecolor{tealblue}{RGB}{20,100,200}

\definecolor{rowblue}{RGB}{220,230,240}

\usepackage{subfiles}

\usepackage{booktabs}
\usepackage{multicol}
\definecolor{rowblue}{RGB}{220,230,240}

\usepackage{capt-of}

\begin{document}
\title{The 2019 DAVIS Challenge on VOS:\\Unsupervised Multi-Object Segmentation}

\author{Sergi Caelles, Jordi Pont-Tuset, Federico Perazzi,\\
        Alberto Montes, Kevis-Kokitsi Maninis, Luc Van Gool~%
\IEEEcompsocitemizethanks{\IEEEcompsocthanksitem S. Caelles, K.-K. Maninis, and L. Van Gool are with the Computer Vision Laboratory, ETH Z\"urich, Switzerland.\protect\\
\IEEEcompsocthanksitem J. Pont-Tuset and A. Montes are with Google AI.\protect\\
\IEEEcompsocthanksitem F. Perazzi is with Adobe Research.\protect\\
}%
\thanks{Contacts and updated information can be found in the challenge website: http://davischallenge.org}}

\markboth{}%
{}

\IEEEtitleabstractindextext{%
\begin{abstract}
We present the \emph{2019 DAVIS Challenge on Video Object Segmentation}, the third edition of the DAVIS Challenge series, a public competition designed for the task of Video Object Segmentation (VOS).
In addition to the original semi-supervised track and the interactive track introduced in the previous edition~\cite{Caelles_arXiv_2018}, a new unsupervised multi-object track will be featured this year.
In the newly introduced track, participants are asked to provide non-overlapping object proposals on each image, along with an identifier linking them between frames (i.e.\ video object proposals), without any test-time human supervision (no scribbles or masks provided on the test video).
In order to do so, we have re-annotated the \texttt{train} and \texttt{val} sets of DAVIS 2017~\cite{Pont-Tuset_arXiv_2017} in a concise way that facilitates the unsupervised track, and created new \texttt{test-dev} and \texttt{test-challenge} sets for the competition.
Definitions, rules, and evaluation metrics for the unsupervised track are described in detail in this paper.
\end{abstract}

}

\maketitle

\IEEEdisplaynontitleabstractindextext

\IEEEpeerreviewmaketitle

\section{Introduction}
The Densely-Annotated VIdeo Segmentation (DAVIS) initiative~\cite{Perazzi2016} supposed a significant increase in the size and quality of the benchmarks for video object segmentation. 
The availability of such dataset introduced the first wave of deep learning based methods in the field~\cite{Caelles2017,Perazzi2017,Cheng2017,Jang2017,Jampani2017,Koh2017,Tokmakov2017,voigtlaender17BMVC,Jain2017,Tokmakov2017a} that considerably improved the state of the art.

The 2017 DAVIS Challenge on Video Object Segmentation~\cite{Pont-Tuset_arXiv_2017} presented an extension of the dataset: 150 sequences (10474 annotated frames) instead of 50 (3455 frames), more than one annotated object per sequence (384 objects instead of 50), and more challenging scenarios such as motion blur, occlusions, etc.
This extended version triggered another wave of deep learning-based methods that pushed the boundaries of video object segmentation not only in terms of accuracy~\cite{Xiao_2018_CVPR, luiten2018premvos, Bao_2018_CVPR, Han_2018_CVPR, Chandra_2018_CVPR, Li_2018, Voigtlaender_2019} but also in terms of computational efficiency, giving rise to fast methods that can even operate online~\cite{Chen1818,Hu_2018_CVPR,Yang_2018_CVPR,Oh_2018_CVPR,Cheng_2018_CVPR,Ci_2018_ECCV,Hu_2018_ECCV,Koh_2018_ECCV,Ventura_2019}.

Motivated by the success of the two previous editions of the challenge, this paper presents the 2019 DAVIS Challenge on Video Object Segmentation, whose results will be presented in a workshop co-located with the Computer Vision and Pattern Recognition (CVPR) conference 2019, in Long Beach, USA. 

We introduce an unsupervised track in order to further cover the spectrum of possible human supervision in video object segmentation at test time: (1) the semi-supervised (introduced in 2017~\cite{Pont-Tuset_arXiv_2017}) track where the masks for the first frame of each sequence are provided, (2) the interactive track (introduced in 2018~\cite{Caelles_arXiv_2018}) where we simulate interactive video object segmentation by using scribble annotations, and (3) the new unsupervised track where methods generate video objects proposals for the entire video sequence without any input at test time.
The main motivation behind the new unsupervised scenario is two-fold.

First, a considerable number of works tackle video object segmentation without human input~\cite{Koh2017, Song_2018_ECCV, Tokmakov2017a, Tokmakov2017, Jain2017, Hu_2018_ECCV, Hu_2018_ECCVb, Li_2018_ECCV, Li_2018_CVPR}. However, in contrast to the semi-supervised scenario, little attention has been given to the case that multiple objects need to be segmented~\cite{Bideau_2018, Xie_2018, Dave_2018}.

Second, annotation of the objects that need to be segmented is a cumbersome process, and the bottleneck in terms of time and effort for very recent methods that solve video object segmentation in real time. Unsupervised methods are able to remove human effort completely, and lead video object segmentation towards fully automatic applications.

We found out that the annotations in DAVIS 2017 are biased towards the semi-supervised scenario, with several semantic inconsistencies. For example, different objects are grouped into a single object in certain sequences whereas the same category of objects are separated in others; or primary objects are not annotated at all.
Even though in the semi-supervised task this does not pose a problem as the definition of what needs to be segmented comes from the mask on the first frame; it would be problematic for unsupervised methods where no information about which objects to segment is provided. To this end, we have re-annotated DAVIS 2017 \texttt{train} and \texttt{val} to be semantically more consistent.

\section{Semi-Supervised Video Object Segmentation}
\textbf{The semi-supervised track remains the same} as in the two previous editions.
The per-pixel segmentation mask for each of the objects of interest is given in the first frame and methods have to predict the segmentation for the subsequent frames. 
We use the same dataset splits:  \texttt{train} (60 sequences),  \texttt{val} (30 sequences),  \texttt{test-dev} (30 sequences), and  \texttt{test-challenge} (30 sequences), 
The evaluation server for  \texttt{test-dev} is always open and accepts an unlimited number of submissions, whereas for  \texttt{test-challenge} submissions are limited in number (5) and time (2 weeks).

The detailed evaluations metrics are available in the 2017 edition manuscript~\cite{Pont-Tuset_arXiv_2017} and more information can be found in the website of the challenge\footnote{\url{https://davischallenge.org/challenge2019/semisupervised.html}}.

\section{Interactive Video Object Segmentation}
\textbf{The interactive track remains the same} as the 2018 edition, except for the use of $\mathcal{J}\&\mathcal{F}$ as the evaluation metric instead of only $\mathcal{J}$ and an \textbf{optional change}: 
In the initial interaction, human-drawn scribbles for the objects of interest in a specific video sequence are provided, and competing methods have to predict a segmentation mask for all the frames in the video and send the results back to a Web Service. The latter simulates a human evaluating the provided segmentation masks and returns extra scribbles in the regions of a frame where the prediction is the worst.
After that, the participant's method refines its segmentation predictions for all the frames taking into account the extra scribbles sent by the Web Server.
This process is repeated several times until a maximum number of interactions (8) or a maximum allowed interaction time (30 seconds per object for each interaction) is reached. 

As an \textbf{optional change} of this year's challenge, methods can also select a list of frames from which the next scribbles will be selected, instead of obtaining the scribbles in the frame with the worst prediction compared to the (unavailable to the public) ground-truth prediction.
In this way, we allow participants to choose the frame(s) for which they need additional information that would improve their results, which is not necessarily the one with the worst accuracy.

In order to allow participants to interact with the Web Server, we use the same Python package that we released in the 2018 edition\footnote{\url{https://interactive.davischallenge.org/}}
More information can be found in the 2018 edition manuscript~\cite{Caelles_arXiv_2018} and in the website of the challenge\footnote{\url{https://davischallenge.org/challenge2019/interactive.html}}.

\section{Unsupervised Video Object Segmentation}

\begin{figure}
\centering
\adjincludegraphics[width=1\linewidth]{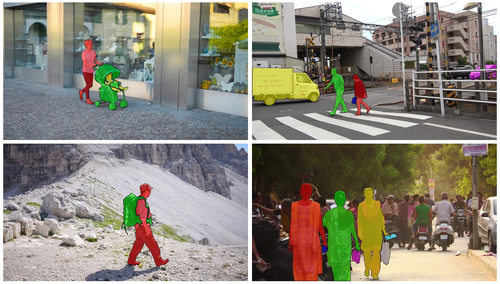}
\caption{\textbf{Re-annotated sequences for the DAVIS 2017 Unsupervised}: Four examples of the re-annotated sequences from the \texttt{train} and \texttt{val} sets of DAVIS 2017 Semi-supervised in order to fulfill the unsupervised definition.}
\label{fig:robot}
\end{figure}

\begin{table*}
\centering
\begin{tabular}{l|ccc|ccccc}
\toprule
 & \multicolumn{3}{c|}{DAVIS 2016} & \multicolumn{5}{c}{DAVIS 2017 Unsupervised} \\
 & \texttt{train} & \texttt{val} & \bf Total & \texttt{train$^{*}$} & \texttt{val$^{*}$} & \texttt{test-dev} & \texttt{test-challenge} & \bf Total\\
 \midrule
 Number of sequences                &	30	& 20	& \bf50 	& 60	& 30	& 30	& 30	&	\bf 150\\
 Number of frames               	& 2079	& 1376	& \bf3455	& 4209	& 1999	& 2294	& 2229	&	\bf 10731\\
Mean number of frames per sequence	& 69.3	& 68.8	& \bf69.1	& 70.2	& 66.6	& 76.46	& 74.3	&	\bf 71.54\\
Number of objects	                & 30    & 20	& \bf50	& 150	& 66	& 115	& 118	&	\bf 449\\
Mean number of objects per sequence	& 1	& 1	& \bf1	& 2.4	& 2.2	& 3.83	& 3.93	&	\bf 2.99\\
\bottomrule
\end{tabular}
\vspace{2mm}
\caption{Size of DAVIS 2017 Unsupervised vs. DAVIS 2016. Sets marked with $^{*}$, contain the same video sequence than the original DAVIS 2017, but they have been re-annotated to fulfill the definition of unsupervised video multi-object segmentation}
\label{tab:sizes}
\vspace{-4mm}
\end{table*}

In the literature, several datasets for unsupervised video object segmentation (in the sense that no human input is provided at test time) have been proposed~\cite{Narayana_2013, Ochs_2014, Bideau2_2016, Galasso_2013, Li_2013, Bideau_2016} often interpreting the task in different ways. 
For example, the authors of the Freiburg-Berkeley Motion Segmentation Dataset~\cite{Ochs_2014} defined the task based on the Gestalt principle of ``common fate''~\cite{koffka2013principles} i.e.\ pixels which share the same motion are grouped together. 
However, this definition does not incorporate any higher understanding of the image and it splits objects that contain different motion patterns. 
In order to account for the latter,~\cite{Bideau_2016} re-annotated the datasets of~\cite{Narayana_2013, Ochs_2014, Bideau2_2016} and re-defined the task to take into account object motion instead of single-pixel motion.
Moreover, objects that are connected in the temporal domain and share the same motion are considered a single instance.

In this work, although we share some similarities with the previous definitions, we give more importance to object semantics rather than their motion pattern.
For example, if a person is holding a bag, we consider the person and the bag as two different objects, even though they may have a similar motion pattern.

In the subsequent sections, we provide a precise definition for unsupervised multi-object video object segmentation, propose an evaluation metric, and describe the rules for this year's track.

\paragraph*{\textbf{Definition}} In the first paper of the DAVIS series~\cite{Perazzi2016}, a single segmentation mask that contains one or more objects joined into a single mask is provided for each frame in a video sequence. In such setting, unsupervised video object segmentation is defined as segmenting all the objects that consistently appear throughout the entire video sequence and have predominant motion in the scene.
However, DAVIS 2017~\cite{Pont-Tuset_arXiv_2017} extended DAVIS 2016 to multiple objects per frame. As a result, the initial definition for unsupervised video object segmentation that considers all objects as one can no longer be used. Thus, we re-define the task by answering two questions: which of the objects that appear in a scene should be segmented and how should objects be grouped together?

In order to define the former, we annotate objects that would mostly capture human attention when watching the whole video sequence i.e\ objects that are more likely to be followed by human gaze. 
Therefore, people in crowds or in the background are not annotated. 
Recently, the same definition was used in~\cite{Wang_2019_CVPR} where the authors recorded eye movement in the DAVIS 2016 sequences and showed that there exist some ‘universally-agreed’ visually important cues that attract human attention.
Moreover, following the definition for all sequences of DAVIS, objects have to consistently appear throughout the video sequence and be present in the first frame.

To answer the second question, we consider that pixels of objects that are enclosed by other objects belong to the latter, for example, people in a bus are annotated as the bus. Furthermore, any object that humans are carrying i.e\ sticks, bags, and backpacks are considered a separate object.

\paragraph*{\textbf{Evaluation metrics}}
In the evaluation, all objects annotated in our ground-truth segmentation masks fulfill the previous definition without any ambiguity.
However, in certain scenarios it is difficult to differentiate between objects that would capture human attention and the ones that would not.
Therefore, predicted segmentation masks for objects that are not annotated in the ground-truth are not penalized.

Methods have to provide \textbf{a pool of $N$ non-overlapping video object proposals for every video sequence} i.e.\ a segmentation mask for each frame in the video sequence where the mask identity for a certain object has to be consistent through the whole sequence.
During evaluation, each of the annotated objects in the ground-truth is matched with one of the $N$ video object proposals predicted by the methods that maximize a certain metric, i.e.,\ $\mathcal{J}\&\mathcal{F}$, using a bipartite graph matching. 

Formally, we have a pool of $N$ predicted video objects proposals $O = \{O_1, ...,O_N\}$ and $L$ annotated objects in the ground-truth $O^{GT} = \{O^{GT}_1, ..., O^{GT}_L\}$. 
We match each object in $O^{GT}$ with only one object in $O$, and every object in $O$ can only be used once.
We define the accuracy matrix $A$ as the result of every possible assignment between $O^{GT}$ and $O$:
\begin{equation*}
    A[l, n] = M(O^{GT}_l, O_n), l \in [1,...,L], n \in [1,...,N]
\end{equation*}
where $A$ has dimensions $L \times N$ and $M$ is the metric that we want to maximize. As in the semi-supervised track, we use the $\mathcal{J}\&\mathcal{F}$ metric~\cite{Pont-Tuset_arXiv_2017}.

We want to find the boolean assignment matrix $X$ where $X[l,n]=1$ iff the predicted object of row $l$ is assigned to the ground-truth object of column $n$. Therefore, we obtain the optimal assignment when:
\begin{equation*}
    X^* = arg\max_X \sum_l\sum_n A_{l,n} X_{l,n}
\end{equation*}
The latter problem is known as maximum weight matching in bipartite graphs and can be solved using the Hungarian algorithm~\cite{Kuhn_1955}.
We use the open source implementation provided by~\cite{scipy}.

For a certain video sequence, the final result will be the accuracy for each ground-truth object taking into account the optimal assignment.
The final performance is the average performance of all objects in the set, exactly as the semi-supervised case.
In this way, we do not weight differently sequences with different number of objects.

We use one of the few available methods that support multi-object unsupervised video object segmentation, RVOS~\cite{Ventura_2019}. We use their publicly available implementation~ \footnote{https://github.com/imatge-upc/rvos} for zero-shot configuration, by generating 20 video object proposals in each sequence. 
In Table~\ref{tab:evaltable2}, we show their performance in the \texttt{val} and \texttt{test-dev} sets. The relatively low performance compared to semi-supervised video object segmentation highlights the intricacy of the unsupervised task.

\begin{table}[t!]
\setlength{\tabcolsep}{4pt} %
\centering
\resizebox{\linewidth}{!}{%
\sisetup{detect-weight=true}
\begin{tabular}{c|c|ccc|ccc}
\toprule
 \multicolumn{8}{c}{DAVIS 2017 Unsupervised}  \\
 \cmidrule(lr){2-8} 
 & $\mathcal{J}$\&$\mathcal{F}$ & $\mathcal{J}$ Mean& $\mathcal{J}$ Recall & $\mathcal{J}$ Decay & $\mathcal{F}$ Mean& $\mathcal{F}$ Recall & $\mathcal{F}$ Decay \\
\cmidrule(lr){1-8}
		\texttt{val}  &	41.2 & 36.8 & 40.2 & 0.5 & 45.7 & 46.4 & 1.7  \\
		\texttt{test-dev}  & 22.5 & 17.7 & 16.2  & 1.6 & 27.3 & 24.8 & 1.8\\
\bottomrule
\end{tabular}
}
\vspace{2mm}
\caption{\label{tab:evaltable2}Performance of RVOS~\cite{Ventura_2019} in the different sets of DAVIS 2017 Unsupervised as a baseline for the newly introduced task.}
\vspace{-2mm}
\end{table}

\paragraph*{\textbf{Challenge}}
The sequences of \texttt{train} and \texttt{val} sets remain the same as the ones from the original DAVIS 2017. However, the masks have been re-labeled in order to be consistent with the aforementioned definitions (the masks for the other tracks are not affected by this modification).
Furthermore, new  \texttt{test-dev} and  \texttt{test-challenge} sets are provided for the unsupervised track.%
 We denote the new annotations as DAVIS 2017 Unsupervised in contrast of the annotations released in the original DAVIS 2017~\cite{Pont-Tuset_arXiv_2017} that we denote as DAVIS 2017 Semi-supervised.
More details can be found in the website of the challenge\footnote{\url{https://davischallenge.org/challenge2019/unsupervised.html}}.

\section{Conclusions}
This paper presents the 2019 DAVIS Challenge on Video Object Segmentation. There will be 3 different tracks available: semi-supervised, interactive, and unsupervised video object segmentation.

In this edition of the challenge, we introduce the new unsupervised multi-object video segmentation track. 
We provide the definition of the task and we re-annotate DAVIS 2017 \texttt{train} and \texttt{val} sets to be consistent with the definition. We additionally introduce 60 new sequences for the \texttt{test-dev} and  \texttt{test-challenge} sets of the unsupervised competition track, and provide the rules and the evaluation metrics for the task.

\ifCLASSOPTIONcompsoc
  \section*{Acknowledgments}
\else
  \section*{Acknowledgment}
\fi

Research partially funded by the workshop sponsors: Google, Adobe, Disney Research, Prof. Luc Van Gool's Computer Vision Lab at ETHZ, and Prof. Fuxin Li's group at the Oregon State University. New unsupervised annotations done by Lucid Nepal.

\ifCLASSOPTIONcaptionsoff
  \newpage
\fi

\bibliographystyle{IEEEtran}

\bibliography{DAVIS2019}

\end{document}